\def\year{2021}\relax
\documentclass[letterpaper]{article} 
\usepackage{aaai21}  
\usepackage{times}  
\usepackage{helvet} 
\usepackage{courier}  
\usepackage[hyphens]{url}  
\usepackage{graphicx} 
\usepackage{natbib}  
\usepackage{caption} 
\usepackage{paralist}
\usepackage{booktabs,multirow}
\usepackage{amsfonts}
\usepackage{amsmath}
\usepackage{bm}
\usepackage{diagbox}
\usepackage{mathrsfs}
\usepackage{algorithm}
\usepackage{algorithmic}
\usepackage{makecell}
\usepackage{diagbox}
\usepackage{multirow}
\usepackage{color, soul}
\usepackage[table,xcdraw]{xcolor}
\usepackage{CJKutf8}

\newcommand{\newcite}[1]{\citeauthor{#1} (\citeyear{#1})}

\begin{document}
 \begin{CJK*}{UTF8}{gbsn}
%
\title{Writing Polishment with Simile: Task, Dataset and A Neural Approach}
\author{
Jiayi Zhang, Zhi Cui, Xiaoqiang Xia, Yalong Guo,\\ Yanran Li, Chen Wei, Jianwei Cui\\}

\affiliations{
Xiaomi AI Lab, Beijing, China\\
\{zhangjiayi3, cuizhi, xiaxiaoqiang, guoyalong, liyanran, weichen, cuijianwei\}@xiaomi.com\\
}

\maketitle
\begin{abstract}
A simile is a figure of speech that directly makes a comparison, showing similarities between two different things, e.g. ``Reading papers can be dull sometimes, \textit{like watching grass grow}". Human writers often \textit{interpolate} appropriate similes into proper locations of the plain text to vivify their writings. However, none of existing work has explored neural simile interpolation, including both locating and generation. In this paper, we propose a new task of Writing Polishment with Simile (WPS) to investigate whether machines are able to polish texts with similes as we human do. Accordingly, we design a two-staged Locate\&Gen model based on transformer architecture. Our model firstly \textit{locates} where the simile interpolation should happen, and then \textit{generates} a location-specific simile. We also release a large-scale Chinese Simile (CS) dataset containing 5 million similes with context. The experimental results demonstrate the feasibility of WPS task and shed light on the future research directions towards better automatic text polishment.

\end{abstract}

\section{Introduction}
Figurative language, or a figure of speech (修辞), is phrasing that goes beyond the literal meaning of words to get a message or point across. Writers and poets use figurative language to build imagery and elicit aesthetic experiences~\cite{citron2018neuroimaging}. 
In computational linguistics, figurative language processing (FLP) has long been an interesting research topic, including both detection \cite{DBLP:conf/naacl/LiS10,DBLP:conf/acl-figlang/2020} and generation tasks \cite{DBLP:conf/emnlp/MishraTS19,DBLP:conf/naacl/Yu019,DBLP:conf/acl/LiuFCMTNZ19}.
\begin{table}[htbp]
\centering
\resizebox{\columnwidth}{!}{
\begin{tabular}{|c|c|l|}
\hline
\rowcolor[HTML]{ECF4FF} 
\multicolumn{3}{|c|}{\cellcolor[HTML]{ECF4FF}Writing Polishment with Simile} \\ \hline
\multicolumn{1}{|c|}{Before} & \multicolumn{2}{l|}{\begin{tabular}[c]{@{}l@{}}Looking at his bloodthirsty eyes, everyone felt\\ horrible and couldn't help but step back.\end{tabular}} \\ \hline
\multicolumn{1}{|c|}{After} & \multicolumn{2}{l|}{\begin{tabular}[c]{@{}l@{}}Looking at his bloodthirsty eyes, everyone felt\\ horrible \underline{\textbf{as if they were being stared at by a }}\\ \underline{\textbf{serpent}}, and couldn't help but step back.\end{tabular}} \\ \hline
\rowcolor[HTML]{ECF4FF} 
\multicolumn{3}{|c|}{\cellcolor[HTML]{ECF4FF}Other Figurative Language Generation} \\ \hline
\rowcolor[HTML]{ECF4FF} 
Task & Status & \multicolumn{1}{c|}{\cellcolor[HTML]{ECF4FF}Text} \\ \hline
 & Before & A metaphorical pair of (Enjoy, Devour) \\ \cline{2-3} 
\multirow{-2}{*}{Metaphor} & After & She devoured his novels. \\ \hline
 & \begin{tabular}[c]{@{}l@{}}Non-\\ ironic\end{tabular} & \begin{tabular}[c]{@{}l@{}}Tried to leave town and my phone died, \\ what a failure.\end{tabular} \\ \cline{2-3} 
\multirow{-2}{*}{\begin{tabular}[c]{@{}c@{}}Irony\end{tabular}} & Ironic & \begin{tabular}[c]{@{}l@{}}\underline{\textbf{Nice}} to leave town and my phone died, \\ \underline{\textbf{definition of success}.}\end{tabular} \\ \hline
\end{tabular}
}
\caption{\label{tab1} Example of WPS and other related tasks.
}
\end{table}

There exist a handful of figurative types that help make concepts become vivid and graspable, including but not limited to simile (明喻), metaphor (隐喻), irony, etc. 
Among them, similes play a vital role for human writings to be attractive. Different from metaphors' using implicit comparisons, a simile is a description that uses ``like'' or ``as'' to make a clear comparison between two separate concepts. As shown in Table \ref{tab1}, human writers add coherent similes into proper locations of the original text to vivify plain writings. Such an \emph{interpolation}-based text polishing process is especially unique for similes, since most polishing objectives clearly requires text rephrasing, e.g., grammar error correction for fluency polishment, text editing for irony style transfer, etc. Distinctly, interpolating similes is like putting proper ingredients to an unflavored dish, instead of totally re-cooking a new one based on a different recipe. Despite the importance of simile, only a few work has explored simile recognition~\cite{DBLP:conf/emnlp/LiuHSFLH18,DBLP:conf/aaai/ZengSSXSL20}. To the best of our knowledge\footnote{We encourage readers to also refer to a contemporary work by \newcite{DBLP:conf/emnlp/ChakrabartyMP20}, which shares a different point of view of simile generation.}, none of existing work has ever investigated simile generation given a plain text, which is indispensable for amplifying writing with similes. 

Although sequence-to-sequence models work well for story generation \cite{DBLP:conf/aaai/LiuLYHL0020}, irony generation \cite{DBLP:journals/corr/abs-1909-06200}, or metaphor and personification generation \cite{DBLP:conf/acl/LiuFCMTNZ19}, it is non-trivial for these models to generate proper and creative simile for a given text. In particular, writing polishment with similes is a unique task because it requires to together address the challenges listed below:
\begin{itemize}
\item Locating is critical for simile interpolation. A simile inserted at a wrong place will impact language fluency and result in weird reading experience. 
\item To polish writing appropriately, the generated simile must be coherent to the context and diverse in the semantics and expressions.
\item Since a simile is disentangled from its context, existing methods are hardly applicable~\cite{DBLP:conf/nips/GuWZ19,DBLP:conf/emnlp/MalmiKRMS19,DBLP:conf/acl/SuSZSHNZ19} since they do not target on interpolation-based text editing.  
\item Like most text style transfer or figurative language generation tasks \cite{DBLP:conf/naacl/RaoT18,DBLP:conf/acl/TsvetkovBSP18,DBLP:journals/corr/abs-1909-06200,DBLP:conf/naacl/Yu019}, there is no large corpus suitable for learning simile interpolation.
\end{itemize}

To this end, we propose a new task of Writing Polishment with Simile (WPS)—to firstly decide \textit{where} to put a simile within plain input text, then figure out \textit{what} content to generate as a coherent simile. To facilitate our research, we propose a new Chinese Simile (CS) dataset, which contains roughly 5.5 million similes in fictional contexts. 
We also set up a benchmark model Locate\&Gen to validate the feasibility and potentials of WPS task. Locate\&Gen model is a two-stage biased generation model upon the framework of transformer encoder-decoder \cite{DBLP:conf/nips/VaswaniSPUJGKP17}. At the first step, it \textit{locates} a pointer position for simile interpolation, and then \textit{generates} a location-specific simile using a novel insertion bias. The two-stage design allows both automatic and semi-automatic inference modes to assist writing polishment flexibly. To summarize, our contributions are three-folded:
\begin{itemize}
\item We introduce the task of Writing Polishment with Simile (WPS), which we believe is a critical step towards figurative writing polishment.
\item We develop\footnote[1]{\url{https://github.com/mrzjy/writing-polishment-with-simile.git}} a large-scale Chinese Simile (CS) dataset for public research, which contains millions of similes with contexts extracted from Chinese online fictions.
\item We establish benchmark model Locate\&Gen and compare it with several SOTA models on the CS dataset, by which we analyze the task in detail.
\end{itemize}

\section{Related Work}
\paragraph{\textbf{Figurative Language Generation}} As a figure of speech, simile generation is related to general figurative language generation.  
\newcite{DBLP:conf/naacl/Yu019} first studied on end-to-end framework for metaphor generation, and \newcite{DBLP:conf/clsw/ZhengSHFZ19} integrated template-based and CBOW-based metaphor generation into chatbots. \newcite{DBLP:conf/emnlp/MishraTS19} proposed multi-staged reinforced seq2seq and retrieval framework for irony generation, while \newcite{DBLP:journals/corr/abs-1909-06200} applied semi-supervised back translation approach with trained irony classifier. 
All these works applied unsupervised learning due to data insufficiency, and none of them considered the scenario of text polishing as WPS. Despite that \newcite{DBLP:conf/acl/LiuFCMTNZ19} explored metaphor and personification over modern Chinese poetry generation, their model is only able to generate new lines instead of polishing existing ones. Similar shortcomings are also observed in most works on story generation~\cite{DBLP:conf/emnlp/TuDYG19,DBLP:conf/acl/FanLD19}. \newcite{DBLP:journals/tacl/GuanHHZZ20} recently devised a knowledge-enhanced model, but it is hard to generate figurative language without explicit goals. Especially for simile, \newcite{DBLP:conf/icccrea/Harmon15} developed a rule-based approach to form a simile given two comparable concepts, and a few work performed simile recognition more recently~\cite{DBLP:conf/emnlp/LiuHSFLH18,DBLP:conf/aaai/ZengSSXSL20}. Our work differs from them in that we aim to polish text with simile interpolation, where context is important for the generated similes to be coherent.
\paragraph{\textbf{Text Style Transfer}} Writing polishment is also related to text style transfer, where the core is the disentanglement of content and implicit attributes called ``style". 
\newcite{DBLP:conf/nips/WangH019} and \newcite{DBLP:conf/iclr/DathathriMLHFMY20} explored plug-and-play schema for controllable text editing. \newcite{DBLP:conf/acl/TsvetkovBSP18} and \newcite{DBLP:conf/nips/YangHDXB18} proposed back-translation and language discriminators respectively to deal with unsupervised text style transfer. Notably, three important differences distinguish our WPS task from previous ones: \textbf{1)} WPS strongly depends on contexts, i.e. the original plain text. \textbf{2)} Unlike the latent ``style", a simile is clearly separable from the semantic content of the original writing. In other words, WPS task requires simile interpolation instead of sentence rephrasing. \textbf{3)} WPS is unique in the need of locating where to insert a simile. Hence, the challenges for WPS include positioning accuracy, contextual coherence and aestheticness of the generated simile.

\paragraph{\textbf{Text Editing at Specific Location}}
Past work designed sequence operations to refine a given piece of text for Grammar Error Correction and Contextual Query Rewrite~\cite{DBLP:conf/acl/SuSZSHNZ19}. \newcite{DBLP:conf/nips/GuWZ19} developed Levenshtein Transformer that supports deletion and insertion operations through iterative decoding schema. \newcite{DBLP:conf/acl/SuSZSHNZ19} applied copy mechanism for transformer encoder-decoder to accomplish utterance rewriting. These methods are designed to process the whole input sequence, which is not suitable for generating similes at specific positions. \newcite{DBLP:conf/emnlp/MalmiKRMS19} developed LASERTAGGER that combines sequence tagging of keep, delete and swap with local sequence generation based on fixed common vocabulary. However, WPS involves only a single simile insertion place, and a common vocabulary would impact simile diversity. In this work, inspired by \newcite{DBLP:conf/acl/LiFMHWL20} who proposed a unified MRC framework for named entity recognition, we adopt a simple but effective approach called Locate\&Gen to realize writing polishment with similes.

\section{Chinese Simile Dataset}
In this section, we present our large-scale Chinese Simile Dataset on its collection, processing as well as qualitative and quantitative analysis.

\paragraph{Data Collection} Although
\begin{table}[ht!]
\resizebox{\columnwidth}{!}{
\begin{tabular}{|c|c|c|c|c|c|}
\hline
\rowcolor[HTML]{ECF4FF} 
\multicolumn{3}{|c|}{\cellcolor[HTML]{ECF4FF}Data Split} & \multicolumn{2}{c|}{\cellcolor[HTML]{ECF4FF}Average Text Length} & \cellcolor[HTML]{ECF4FF} \\ \cline{1-5}
\cellcolor[HTML]{EFEFEF}Train & \cellcolor[HTML]{EFEFEF}Dev & \cellcolor[HTML]{EFEFEF}Test & \cellcolor[HTML]{EFEFEF}Context & \cellcolor[HTML]{EFEFEF}Simile & \multirow{-2}{*}{\cellcolor[HTML]{ECF4FF}\begin{tabular}[c]{@{}c@{}}\# Unique\\ Simile\end{tabular}} \\ \hline
5,485,721 & 2,500 & 2,500 & 52.7 & 8.3 & 3,040,544 \\ \hline
\rowcolor[HTML]{ECF4FF} 
\multicolumn{3}{|c|}{\cellcolor[HTML]{ECF4FF}Simile Position Distribution} & \multicolumn{3}{|c|}{\cellcolor[HTML]{ECF4FF}Simile Complexity Distribution} \\ \hline
\rowcolor[HTML]{ECF4FF} 
\cellcolor[HTML]{EFEFEF}Start & \cellcolor[HTML]{EFEFEF}Middle & \cellcolor[HTML]{EFEFEF}End & \cellcolor[HTML]{EFEFEF}Single & \cellcolor[HTML]{EFEFEF}Combined & \cellcolor[HTML]{EFEFEF}Clause \\ \hline
4.2\% & 82.3\% & 13.5\% & 37\% & 29\% & 34\% \\ \hline
\end{tabular}
}
\caption{\label{tab:stat} Statistics and properties of Chinese Simile Dataset.}
\end{table}
\begin{table}[ht!]
\small
\resizebox{\columnwidth}{!}{
\begin{tabular}{|c|c|}
\hline
\multicolumn{2}{|c|}{\cellcolor[HTML]{ECF4FF}Samples from CS dataset} \\ \hline
\multirow{4}{*}{\#1} &  \multicolumn{1}{l|}{\begin{minipage}[t]{0.9\columnwidth} %
他\underline{\textbf{像幽灵一样}}出现那里，单手握门框上，挡住了那女的退路。耀眼如银丝般的长发下... \\
He appeared there \underline{\textbf{like a ghost}}, holding the door frame with one hand, blocking the her retreat...%
\end{minipage}} \\ \hline
\multirow{5}{*}{\#2} & \multicolumn{1}{l|}{\begin{minipage}[t]{0.9\columnwidth} %
那些狼首怪物休想靠近她，一旦接近十米的范围，就\underline{\textbf{像是被流沙沼泽吞噬掉似的}}陷入地下。\\
Those wolf-head monsters ... would suddenly sink into the ground \underline{\textbf{as if swallowed by a quicksand swamp}} when getting close to her in 10 meters range.%
\end{minipage}} \\ \hline
\end{tabular}
}
\caption{\label{tab:sample} Samples from Chinese Simile Dataset. The extracted ground-truth similes within context are underlined. Translation is provided for non-Chinese speakers.}
\end{table}
\newcite{DBLP:conf/emnlp/LiuHSFLH18} has introduced a simile recognition dataset based on student essays, it only contains 5,088 simile samples extracted from student essays, which is insufficient for generating diverse and high-quality similes. In comparison, \newcite{DBLP:conf/emnlp/ChakrabartyMP20} collected comments containing similes from Reddit and then auto-constructed a parallel simile corpus thanks to the commonsense knowledge inference by COMET \cite{DBLP:conf/acl/BosselutRSMCC19}, and finally obtained 80k+ self-labeled human written similes. However, their similes are all limited in a ``like a" pattern and appear only at the end of a sentence, which is still different from real-world setting. Instead, we choose as our data sources the online free-access fictions that are tagged with sci-fi, urban novel, love story, youth, etc, because fiction writers often use figurative language to engage their audience. 
The raw corpus we accessed and crawled contains 22,463 online Chinese fictions for 60GB of text with roughly 470 million characters. 

\paragraph{Data Processing} We split long paragraphs into pieces of sentences and re-combine them to our contextualized simile samples, whose max length is set to 128. Thus, each sample in our CS dataset typically corresponds to several continuous sentences or a whole paragraph from fictions. We automatically extract sentences that match a dozen of rich Chinese simile start and ending patterns (e.g. ``好像", ``仿佛", ``宛若", ``俨然", ``如同", ``似的", ``一般", etc., all meaning ``as if" or ``like"). Similes containing personage names are eliminated (e.g. ``as beautiful as Elizabeth") by Jieba's name recognition\footnote[1]{https://github.com/fxsjy/jieba}. For similes that occur more than 100 times, we downsample them by a keep ratio of 100/occurrence so as to mitigate the problem of generating frequent but meaningless similes. Detailed statistics could be found in Table~\ref{tab:stat}.

\paragraph{Data Analysis} We ask three annotators to assess the properties of 500 random samples from CS dataset on multiple aspects, and present the assessment in Table~\ref{tab:stat}.
\begin{itemize}
\item \textbf{Data Quality.} Although \newcite{DBLP:conf/emnlp/LiuHSFLH18} claimed that neural methods lead to higher recognition accuracy than rule-based ones, all their simile data is constructed using a single pattern, i.e., sentences containing the comparator ``像". In contrast, we adopt more complex regex patterns and ensure an extraction precision of 92\% approximately. Incorrect extractions are either due to over-recall on simile patterns or failures of Jieba's name recognition. Given the massive unique similes, we didn't pursue recalling more similes at the risk of hurting precision. 
\item \textbf{Variety of Simile Position.} We investigate whether the simile appears at the start, middle or end of the entire content. It turns out that 82.3\% of similes are at somewhere middle of a text, demonstrating the strong feature of interpolation.
\item \textbf{Generation Challenge.} To understand the generation difficulty of the task, We distinguish the simile complexity on three levels. Ranging from simple to hard, we define similes containing only a noun or a verb as single, ``ADJ. + noun" or ``ADV. + verb" as combined, and attributive clause as clause. The statistics show that a decent amount of similes in fictions include complex expressions, such as sample \#2 shown in Table~\ref{tab:sample}. Drawing on the overall analysis, we find that human writers interpolate similes at various positions in different shapes. This implies that for WPS task, simile diversity should be also considered during evaluation. 
\end{itemize}
\begin{figure*}[ht]
\centering
\resizebox{0.9\linewidth}{!}{
    \includegraphics{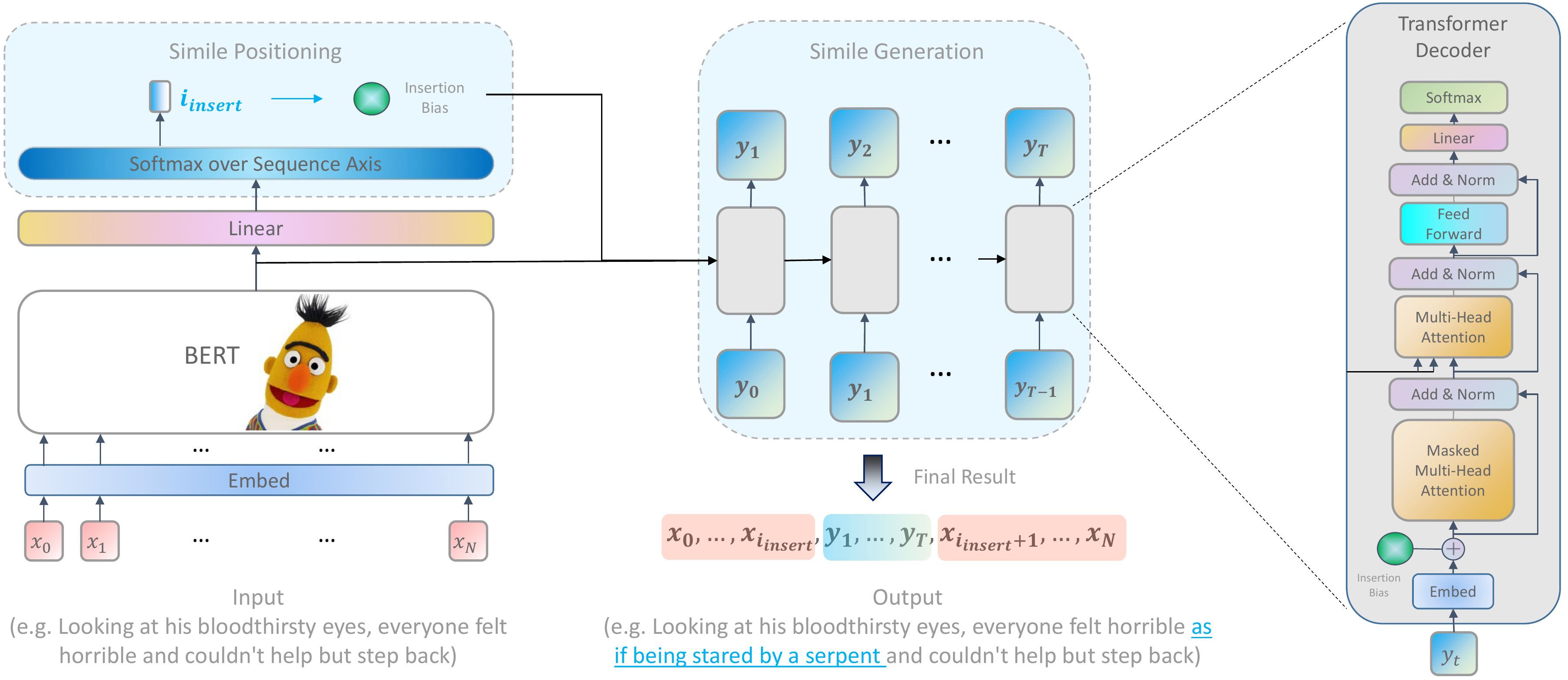}
}
\caption{\label{fig:1}Illustration of Locate\&Gen model that is derived from transformer encoder-decoder framework. The first stage of simile positioning is shown at the upper left, the generation stage is at the right. An insertion vector is obtained based on positioning result that bias the decoder towards desired sequence generation.}
\end{figure*}

\section{Model}
\subsection{Task Definition and Overview}
The task of Writing Polishment with Simile (WPS) is formulated as follows: Given a sequence of context tokens $X=\{x_0, ..., x_{N}\}$, the goal is to insert a coherent simile $Y=\{y_1, ..., y_{T}\}$ at a proper position of $X$. In order to mimic human behaviors, we cast this task as a two-staged process: The first stage is simile positioning, which predicts where to insert a simile within $X$. The insertion position is denoted as $i_{\text{Ins}}$, where $0\leq i_{\text{Ins}}\leq N$. The second stage is simile generation, which generates a simile $Y$ word-by-word considering the context $X$ and the predicted $i_{\text{Ins}}$. Hence, the probability of generating a simile $Y$ can be depicted as:
\begin{equation} 
\small
\begin{split}
P_Y = \prod_{i=1}^{T}P(y_t|y_{0:t-1}, i_{\text{Ins}}, X)
\end{split}
\label{eq:overall}
\end{equation}
\subsection{Locate\&Gen Architecture}
\paragraph{\textbf{Encoder}} As shown in Figure~\ref{fig:1}, we design Locate\&Gen framework. Typically, we apply BERT~\cite{DBLP:conf/naacl/DevlinCLT19} as our encoder. It takes $X$ as input and encode it into hidden vector sequence $\textbf{h}=\{h_0, h_1, ..., h_{N}\}$ based on multi-head self-attention, where $h\in \mathbb{R}^H$ and $H$ is the hidden size. This process is depicted as follows:
\begin{equation} 
\small
\begin{split}
\textbf{e} & = \{\text{Embed}(x_0), ..., \text{Embed}(x_{N})\} \\
\textbf{h} & = \text{BERT}(\textbf{e})
\end{split}
\label{eq:encoder}
\end{equation}
where $Embed$ is the embedding function (e.g., the sum of word, segment and position embedding) that transforms a token $x_i$ into a vector $e_i\in \mathbb{R}^E$. Note that dimension $E$ is set to be equal to hidden size $H$ for standard transformer model.
\paragraph{\textbf{Simile Positioning}} To compute the probability of a position $i\in [0,N]$ being the insertion position pointer $i_{\text{Ins}}$, we apply softmax over all of projected hidden vectors along sequence axis:
\begin{equation}
\small
P^{\text{Ins}}_i = \frac{\exp(h_iW_{\text{Ins}})}{\sum^N_j \exp(h_jW_{\text{Ins}})}, i\in [0, N]
\label{eq:pointer}
\end{equation}
where $W_{\text{Ins}} \in \mathbb{R}^{H \times 1}$ is a pointer weight matrix, and $h_i$ is the i-th hidden vector of encoder outputs $\textbf{h}$ according to equation \ref{eq:encoder}. Note that a special token ``[CLS]" (denoted as $x_0$) is prepended to each input sequence during BERT pretraining phase, which could also be used as a null pointer for cases when no possible insertion location exists. The insertion position $i_{\text{Ins}}$ is simply calculated as follows:
\begin{equation}
\small
i_{\text{Ins}} = \mathop{\mathrm{argmax}}_i (\{P^{\text{Ins}}_i|i\in [0, N]\})
\label{eq:pointer_insert}
\end{equation}
\paragraph{\textbf{Simile Generation}} The decoder is adapted from standard transformer decoder. The decoding process involves not only the self attention among previously decoded tokens (a.k.a. causal attention), but also the encoder-to-decoder attention as well. It takes the encoder output $\textbf{h}$ and the previously decoded hidden vectors $\{s_0, ..., s_{t-1}\}$ as input, and autoregressively produce one  token $y_t$ at each time step. 
\paragraph{\textbf{Insertion Bias}} In order to guide the decoder generation with the signal of simile positioning, we compute an insertion bias vector $k\in R^H$ by projecting the $i_{\text{Ins}}$-th encoder hidden vector as follows:
\begin{equation}
\small
\label{eq:insertion_vec}
k = h_{i_{\text{Ins}}}W_{\text{IB}}
\end{equation}
where $W_{\text{IB}}\in R^{H\times H}$ is a weight matrix. 
By considering the insertion bias, the probability $P_{y_t}$ of generating $y_t$ at each step is finally modeled as:
\begin{equation}
\small
\begin{split}
e_{t-1} & = \text{Embed}(y_{t-1}) + k \\
s_{t} & = \text{TransformerBlock}(\textbf{h}, s_{0:t-1}, e_{t-1}) \\
P_{y_t} & = \text{softmax}(s_tW_e)
\end{split}
\label{eq:decoder}
\end{equation}
where $y_0$ is a special start token, $W_e\in R^{E\times |V|}$ is the word embedding matrix with vocabulary size of $|V|$. The final selection of token $y_t$ could be processed based on auxiliary decoding strategies such as greedy or sampling methods.

\subsection{Training and Inference}
\paragraph{\textbf{Loss}} Since equation \ref{eq:pointer_insert} is not differentiable, we treat the training of positioning and generation as multi-task learning, and minimize the cross-entropy loss respectively for the optimization of the two-staged Locate\&Gen model. The loss is calculated as:
\begin{equation}
\small
\mathcal{L}_{\text{Total}} = \underbrace{(-\sum^N_{i=0} i_{\text{Ins}} \log P^{\text{Ins}}_i)}_{\text{Positioning Loss}} +  \underbrace{(-\sum^T_{t=0} y_t \log P_{\hat{y_t}})}_{\text{Generation Loss}} 
\label{eq:loss}
\end{equation}
where $i_{\text{Ins}}$ and $y_t$ here denote the ground-truth insertion position and gold simile token at $t$-th time step, respectively.
\paragraph{\textbf{Teacher Forcing}} Similar to previous approaches, there also exists an exposure bias in Locate\&Gen framework since during training, we expose ground-truth target, i.e., the real pointer position and gold previous sequence tokens. However, we only observed limited improvement by using techniques such as scheduled sampling \cite{DBLP:conf/nips/BengioVJS15} or Gumbel Softmax Approximations \cite{DBLP:conf/iclr/JangGP17}. The reasons why exposure bias is not that harmful in our case might be several-folds. For one cause, simile is relatively easy to locate and the simile length is often shorter than normal text generation tasks. Also, the strong capacity of the BERT encoder also results in robust performance \cite{DBLP:conf/acl/HsiehCJWHH19}, which help alleviate the issue of exposure bias~\cite{DBLP:journals/corr/abs-2006-10369}. We leave the usage of more complex training techniques to future works.

\paragraph{\textbf{Inference}} Since WPS is interpolation-based, two inference modes can be adopted. \textbf{1) Automatic Mode.} Ideally, our Locate\&Gen model is able to complete the simile positioning and generation given a plain text in a fully automatic way, without the need of explicitly telling the model where to add similes. \textbf{2) Semi-automatic Mode.} Meanwhile, thanks to the design of insertion bias, it's also possible to put simile at any arbitrary location other than the most probable one. In such cases, the model takes an additional input of $i_{\text{Ins}}$ so as to directly start the biased generation. We investigate both of these two modes in the experiments. 

\section{Experiments}
In this section, we benchmark and analyze the performances of baseline models including Locate\&Gen as well as retrieval approaches.

\subsection{Baseline Models}
\paragraph{\textbf{Locate\&Gen}} We set up Locate\&Gen model as introduced in the previous section. To investigate the significance of each component, we also compare with several variants: Locate\&Gen (\textbf{+ beam 20}) is an enhanced version where we apply beam search decoding with beam size of 20. Locate\&Gen (\textbf{- w/o Pretrain}) is an ablated model, in which the BERT encoder is randomly initialized and trained from scratch. 
Locate\&Gen (\textbf{- w/o InsBias}) is another ablated version where we eliminate the insertion bias from the decoder and retrain the model.

\paragraph{Retrieve\&Match} It also makes sense to select proper simile(s) using retrieve-then-rerank techniques. \textbf{1) Retrieve.} Our assumption is that a simile coherent to a given context is also likely to be suitable for similar contexts. In the first step, we treat as context the 16 surrounding characters around the insertion position and retrieve 100 most similar contexts from training corpus using \textbf{BM25} ranking scores \cite{robertson2009probabilistic}. Note that setting the retrieval context length to 16 is an empirical choice, since either shorter or longer context ends in inferior retrieval performance. If no further step exists, the simile with the highest BM25 score is chosen as the model output. \textbf{2) Match.} To improve the retrieval performance, we also collect the top 100 similes from the first step as candidates, and acquire two matching models to re-rank them. \textbf{MatchCBOW} calculates cosine similarity of sentence embeddings (i.e., average of word embeddings) between simile and the original writing. We also adopt \textbf{MatchBERT}, a BERT matching model whose input is the concatenation of context and simile. The simile with the highest matching score is returned as the final output. For fair comparison, all retrieval-based methods share the position predictions from Locate\&Gen model.

\begin{table*}[ht]
\centering
\small
\resizebox{\linewidth}{!}{
\begin{tabular}{|l|c|c|c|c|c|c|c|c|c|c|c|c|}
\hline
\rowcolor[HTML]{ECF4FF} 
\multicolumn{1}{|c|}{\cellcolor[HTML]{ECF4FF}} & \multicolumn{1}{c|}{\cellcolor[HTML]{ECF4FF}Positioning} & \multicolumn{11}{c|}{\cellcolor[HTML]{ECF4FF}Simile Generation} \\ \cline{2-13} 
\rowcolor[HTML]{ECF4FF} 
\multicolumn{1}{|c|}{\cellcolor[HTML]{ECF4FF}} & & \cellcolor[HTML]{ECF4FF} & \multicolumn{3}{c|}{\cellcolor[HTML]{ECF4FF}Word Overlap (With G.T.)} & \cellcolor[HTML]{ECF4FF} & \multicolumn{3}{c|}{\cellcolor[HTML]{ECF4FF}Contextual Similarity} & \multicolumn{3}{c|}{\cellcolor[HTML]{ECF4FF}Diversity} \\ \cline{4-6} \cline{8-13} 
\rowcolor[HTML]{ECF4FF} 
\multicolumn{1}{|c|}{\multirow{-3}{*}{\cellcolor[HTML]{ECF4FF}Model}} & \multirow{-2}{*}{\cellcolor[HTML]{ECF4FF}Accuracy} & \multirow{-2}{*}{\cellcolor[HTML]{ECF4FF}PPL} & BLEU1 & BLEU2 & BLEU3 & \multirow{-2}{*}{\cellcolor[HTML]{ECF4FF}Length} & \cellcolor[HTML]{ECF4FF}EA & \cellcolor[HTML]{ECF4FF}GM & \cellcolor[HTML]{ECF4FF}VE & Dist-1 & \cellcolor[HTML]{ECF4FF}Dist-2 & Dist-S \\ \hline
\multicolumn{1}{|c|}{BM25} & \multicolumn{2}{l|}{\cellcolor[HTML]{ECF4FF}} & 20.00 & 1.37 & 0.25 & 8.6 & 0.777 & 0.177 & 0.540 & 0.088 & 0.472 & \textbf{0.996} \\ \cline{1-1} \cline{4-13}
MatchCBOW & \multicolumn{2}{l|}{\cellcolor[HTML]{ECF4FF}} & 25.13	& 2.96 & 0.56 & \textbf{14.4} & \textbf{0.889} & 0.168 & \textbf{0.608} & 0.060 & 0.402 & 0.988 \\ \cline{1-1} \cline{4-13}
MatchBERT & \multicolumn{2}{l|}{\cellcolor[HTML]{ECF4FF}} & 25.55 & 4.45 & 1.14 & 9.6 & 0.800 & 0.179 & 0.561 & 0.088 & \textbf{0.494} & 0.990 \\ \hline
Locate\&Gen & & & 39.73 & 18.82 & 7.68 & 7.9 & 0.791 & 0.188 & 0.560 & 0.059 & 0.247 & 0.887 \\ \cline{1-1} \cline{4-13}
+ beam 20 & \multirow{-2}{*}{\textbf{0.769}} & \multirow{-2}{*}{\textbf{6.469}} & \textbf{41.28} & \textbf{19.77} & \textbf{8.32} & 7.2 & 0.777 & \textbf{0.190} & 0.553 & 0.064 & 0.270 & 0.855 \\ \hline
- w/o Pretrain & 0.713 & 7.344 & 36.36 & 16.25 & 5.75 & 8.0 & 0.792 & 0.185 & 0.559 & 0.051 & 0.207 & 0.818 \\ \hline
- w/o InsBias & 0.726 & 6.870  & 36.98 & 16.42 & 5.85 & 8.1 & 0.791 & 0.187 & 0.559 & 0.053 & 0.217 & 0.831 \\ \hline

\multicolumn{1}{|c|}{G.T.} & \multicolumn{5}{l|}{\cellcolor[HTML]{ECF4FF}} & 8.3 & 0.782 & 0.183 & 0.552 & \textbf{0.090} & 0.473 & \textbf{0.996} \\ \hline
\end{tabular}
}
\caption{\label{tab:automatic}Evaluation on automatic metrics of baselines on Simile Polishment. Highest scores are in bold. Metric scores of ground-truth (G.T. for short) similes are shown as well. Note that no positioning accuracy is reported for retrieval methods since they share the same simile position predictions from Locate\&Gen.}
\end{table*}

\subsection{Evaluation Metrics}
\paragraph{Positioning Accuracy} We evaluate the simile insertion accuracy compared with ground-truth simile position. The prediction is scored as correct only when it exactly matches the ground-truth location. Note that 
insertions located elsewhere do not necessarily lead to a bad simile interpolation.

\paragraph{Word Overlap (BLEU)} BLEU\footnote[1]{We use the public multi-bleu.perl script for BLEU calculation.} \cite{papineni2002bleu} reflects the word overlap between the generated and the ground-truth text. Typically, we use BLEU$_{n}$ ($n=1,2,3$) considering the n-gram overlaps of similes.

\paragraph{Embedding Similarity} To assess similes' contextual relevance, we use pretrained embeddings\footnote[2]{https://ai.tencent.com/ailab/nlp/en/embedding.html} to calculate embedding average (EA), greedy match (GM) and vector extrema (VE) \cite{DBLP:conf/emnlp/LiuLSNCP16} between simile and context.

\paragraph{Perplexity, Diversity, Length} Perplexity (PPL.) measures how well a language model estimates the probability distribution over entire sentences \cite{DBLP:journals/corr/abs-2001-09977} (lower is better). Meanwhile, diversity computes the ratios of distinct unigrams, bigrams and sentences, denoted as Dist-1, Dist-2 and Dist-S, respectively. Besides, we adopt the generation length as a simple indicator of content richness.

\paragraph{Human Evaluation} We conduct human evaluation in order to assess the overall performance of WPS, namely the simile positioning as well as generation. We adopt in total 4 aspects for human judgments. Fluency (Flu.) and creativity (Crea.) \cite{DBLP:conf/naacl/Yu019} are to examine grammatical correctness of a simile given the surrounding context, and whether it's interesting and creative instead of dull or bland, respectively. In addition to Crea., we adopt informativeness (Info.) to investigate whether a simile is rich in content. Note that a creative simile does not need to be really rich in content, meanwhile a content-rich simile might still be universal or dull. Besides, we propose coherence (Coh.) to assess whether a simile coherently fits its context on semantic aspect. We ask 5 well-educated annotators to assess 100 random samples from test set and rate each simile with \{0, 1\} score on the 4 aspects.

\subsection{Implementation Details}
All models are implemented in Tensorflow\footnote[3]{https://www.tensorflow.org} and trained on Nvidia Tesla V100 GPU. We apply standard BERT-base\footnote[4]{We adopt pretrained Chinese BERT-wwm-ext checkpoint from https://github.com/ymcui/Chinese-BERT-wwm.} settings as the encoder. For the decoder, we set as 2-layer transformer of 768 hidden size. We use standard BERT tokenizer to perform char-level Chinese tokenization with a vocab size of 21,128. During training, we apply embedding weight-tying as well as label smoothing technique~\cite{muller2019does} to improve training speed and robustness. We use dropout of 0.1, and Adam optimizer~\cite{DBLP:journals/corr/KingmaB14} with a mini-batch of 128. The max context and target length are set to 128 and 16 respectively. For MatchBERT, we train with random negative sampling of size 5. Without hyper-parameter tuning, we set the learning rate to 5e-5 and train for maximum of 15 epochs with early stopping on Dev set. 

\begin{table}[t]
\centering
\small
\begin{tabular}{|c|c|c|c|c|}
\hline
\rowcolor[HTML]{ECF4FF} 
\multicolumn{1}{|c|}{\cellcolor[HTML]{ECF4FF}Model} & Flu. & Crea. & Coh. & Info. \\ \hline
BM25 & 0.89 & 0.55 & 0.54 & 0.64 \\ \hline
MatchCBOW & 0.80 & 0.56 & 0.43 & \textbf{0.83} \\ \hline
MatchBERT & 0.96 & 0.59 & 0.64 & 0.79 \\ \hline
Locate\&Gen & 0.95 & 0.54 & 0.77 & 0.58 \\ \hline
\multicolumn{1}{|c|}{\cellcolor[HTML]{EFEFEF}GroundTruth} & \multicolumn{1}{c|}{\cellcolor[HTML]{EFEFEF}\textbf{0.99}} & \multicolumn{1}{c|}{\cellcolor[HTML]{EFEFEF}\textbf{0.66}} & \multicolumn{1}{c|}{\cellcolor[HTML]{EFEFEF}\textbf{0.85}} & \multicolumn{1}{c|}{\cellcolor[HTML]{EFEFEF}0.61} \\ \hline
\end{tabular}
\caption{\label{tab:human}Human evaluation results. The human agreement is validated by Kappa coefficient~\cite{Fleiss1971Measuring} of 0.45, indicating a ``moderate agreement" among annotators.}
\end{table}

\subsection{Results and Analysis}
\subsubsection{Overall Analysis} We first examine the results based on automatic metrics. At a glimpse of Table \ref{tab:automatic}, the best scores for positioning and simile generation are promising but imperfect, suggesting the feasibility and potential of our WPS task. 
When we examine the contextual similarity and diversity scores (the last six columns), things become interesting. The similarity scores for ground-truth similes are comparable and even lower than the best scores achieved by benchmark models. It indicates that contextual similarity is a necessary but not sufficient measurement for WPS task. Moreover, the three similarity metrics favour different models due to different calculation manners. Both EA and VE compute sentence-level embeddings and thus models outputting long similes will gain higher EA and VE scores. It is because long similes often share more similar words with context. GM instead prefers similes resembling the key words in the contexts~\cite{DBLP:conf/emnlp/LiuLSNCP16}. Thus, GM is a more reasonable indicator of contextual similarity especially for noisy contexts. In terms of diversity, there is a clear gap between generated and human-written similes. Overall speaking, similes produced by retrieval-based methods are often more diverse but less coherent than generation-based ones, as implied by their higher Dist-n but poorer BLEU and GM scores.


\begin{table*}[ht]
\centering
\resizebox{\linewidth}{!}{
\begin{tabular}{|c|c|l|l|l|}
\hline
\multicolumn{1}{|c|}{\cellcolor[HTML]{ECF4FF}Case} &
\multicolumn{1}{|c|}{\cellcolor[HTML]{ECF4FF}Model} &  \multicolumn{2}{|c|}{\cellcolor[HTML]{ECF4FF}Simile Generation (with Human Translation)} & \multicolumn{1}{|c|}{\cellcolor[HTML]{ECF4FF}Original Writing} \\ \hline
& \multicolumn{1}{|c|}{\cellcolor[HTML]{EFEFEF}GroundTruth} & \multicolumn{1}{|l|}{\cellcolor[HTML]{EFEFEF}就好像一个倒扣的锅盖一样} & \multicolumn{1}{|l|}{\cellcolor[HTML]{EFEFEF}like an upside-down lid} & \multirow{5}{*}{\begin{minipage}[t]{1.3\columnwidth} %
只见他此刻的位置赫然是一个地下的大空洞。通体呈不规则的圆形，\colorbox{yellow}{[Insert]}，这空洞巨大无比，也空旷的骇人。只有一条地下河从空洞的中央穿过。他就是从那里逃出来的。\\
He was standing right beside a giant irregularly round cavity \colorbox{yellow}{[Insert]}, only an underground river passes through its center. He just escaped from there.%
\end{minipage}} \\ \cline{2-4}
& BM25 & 好像一个巨大的山洞 & like a huge cave & \\ \cline{2-4}
\#1 & MatchCBOW & 好像一个巨大的山洞 & like a huge cave & \\ \cline{2-4}
 & MatchBERT & 如同在溶洞之中一般 & as if in a karst cave & \\ \cline{2-4}
& Locate\&Gen & 就像是一个巨大的漏斗一般 & like a huge funnel & \\ \hline \hline
& \multicolumn{1}{|l|}{\cellcolor[HTML]{EFEFEF}GroundTruth} & \multicolumn{1}{|l|}{\cellcolor[HTML]{EFEFEF}就仿佛被定住了身形一般} & \multicolumn{1}{|l|}{\cellcolor[HTML]{EFEFEF}like being frozen} &
\multirow{5}{*}{\begin{minipage}[t]{1.3\columnwidth}%
而紫僵的动作，戛然而止，\colorbox{yellow}{[Insert]}。忽然就停顿在原地不再动弹。\\
And Zi Jiang's movements stopped abruptly \colorbox{yellow}{[Insert]}. She just paused and did not react any more. %
\end{minipage}} \\ \cline{2-4}
& BM25 & 似木偶一般 & like a wooden puppet & \\ \cline{2-4}
\#2 & MatchCBOW & 就像一个人被猛的掐断了脖子似的 & like a person who has his neck severed & \\ \cline{2-4}
 & MatchBERT & 像是被摁了禁止键似的 & as if someone pressed her stop button & \\ \cline{2-4}
& Locate\&Gen & 就像是被施了定身法一样 & like being casted a freezing charm & \\ \hline \hline
& \multicolumn{1}{|l|}{\cellcolor[HTML]{EFEFEF}GroundTruth} & \multicolumn{1}{|l|}{\cellcolor[HTML]{EFEFEF}像受惊的鹌鹑一样} & \multicolumn{1}{|l|}{\cellcolor[HTML]{EFEFEF}like frightened quails} &
\multirow{5}{*}{\begin{minipage}[t]{1.3\columnwidth}%
几个被扇飞开去的泰坦人这才反应过来，做出戒备的姿态。但看到满脸怒容的泰坦之王，这些泰坦人顿时\colorbox{yellow}{[Insert]}匍匐在地。\\
The few Titans who were knocked in the air finally came to their senses, and turned into vigilant state. However they suddenly prostrated \colorbox{yellow}{[Insert]} after seeing their king glaring at them. %
\end{minipage}} \\ \cline{2-4}
& BM25 & 像是一摊烂泥一样的 & like a mess of mud & \\ \cline{2-4}
\#3 & MatchCBOW & 似乎是被扳倒了一样 & like being knocked down & \\ \cline{2-4}
 & MatchBERT & 如同膜拜神一样 & like worshiping god & \\ \cline{2-4}
& Locate\&Gen & 如同见到了猫的老鼠一般 & like mice confronting a cat & \\ \hline \hline
& \multicolumn{1}{|l|}{\cellcolor[HTML]{EFEFEF}GroundTruth} & \multicolumn{1}{|l|}{\cellcolor[HTML]{EFEFEF}就像个受了委屈的小媳妇般} & \multicolumn{1}{|l|}{\cellcolor[HTML]{EFEFEF}like a wronged little daughter-in-law} &
\multirow{5}{*}{\begin{minipage}[t]{1.3\columnwidth}%
苏乐遥不言不语的样子看上去\colorbox{yellow}{[Insert]}可怜兮兮，又无从说起。\\
Leyao Su remained silent and looked pitiful \colorbox{yellow}{[Insert]}, but she could not actually talk about it.%
\end{minipage}} \\ \cline{2-4}
& BM25 & 就像一只要被带走的大狗般 & like a big dog about to be taken away & \\ \cline{2-4}
\#4 & MatchCBOW & 就好像是受了委屈的小媳妇一样 & like a wronged little daughter-in-law & \\ \cline{2-4}
 & MatchBERT & 像一只害怕被主人抛弃的小狗似的 & like a puppy afraid of being abandoned & \\ \cline{2-4}
& Locate\&Gen & 就像是被抛弃的小狗一样 & like an abandoned puppy & \\ \hline
\end{tabular}
}
\caption{\label{tab:compare}Model comparison of simile generation. For fair comparison, we choose cases where model prediction of simile position exactly matches ground-truth position (noted as [insert] in yellow). We make great effort in faithfully translating the original Chinese narrative writings to English, however there still might be language gaps between them.}
\end{table*}

\begin{table}[ht]
\small
\resizebox{\columnwidth}{!}{
\begin{tabular}{|l|}
\hline
\rowcolor[HTML]{ECF4FF} 
\multicolumn{1}{|c|}{\cellcolor[HTML]{ECF4FF}Automatic / Semi-Automatic Polishment} \\ \hline
\begin{minipage}[t]{\columnwidth}%
梁师这一脚好生了得，王宇感觉\colorbox{yellow}{像是被火车撞了一样}难受，身体就要不顾自己的控制\colorbox{orange}{像断线的风筝一样}飞出悬崖，还好他及时用双手抓在水泥地上，只 \colorbox{orange}{如同鹰爪一样}抓划着地面发出了令人发指的声音。
\\
Wang Yu was kicked so hard by Liang's feet and felt uncomfortable \colorbox{yellow}{like being hit by a train}, he almost seemed to be kicked off the cliff \colorbox{orange}{like a broken kite}. Fortunately, he grabbed the concrete floor \colorbox{orange}{as eagle claws} in time, with ground-scratching sound. %
\end{minipage} \\ \hline
\end{tabular}
}
\caption{\label{case:mode} Polishing modes. Generated similes in automatic and semi-automatic mode are in yellow and orange, respectively. We apply beam search decoding for reproducibility.} 
\end{table}

\subsubsection{Human Assessment} In Table \ref{tab:human}, we observe the human judgments are consistent with automatic evaluations that generative methods have advantages over retrieval ones in terms of coherence, which we assume is a critical and indispensable goal of WPS task. 
Surprisingly, the similes selected by BM25 and MatchCBOW are annotated with lower Flu. scores, even though they are human-written sentences by nature. We conjecture that these methods are prone to focus on context noise during context-simile semantic matching, hence the selected similes even break the overall language fluency when considering the context together. As such, MatchBERT is more capable of distinguishing the features and noise, which yields better performance. Note that even ground-truth simile is not always perfect, and its creativity score of 0.66 and informativeness of 0.61 partially reveal the need of automatic polishing to help humans with their writings.

\subsection{Ablation Study} 
We investigate the impact of beam search (versus beam 20), BERT pretraining (versus w/o Pretrain) as well as insertion bias (versus w/o InsBias) on WPS task. 

\paragraph{Beam Search} Changing greedy decoding to beam search (size=20) results in gains on BLEU scores but losses on length and Dist-S. It is a common trade-off between blandness and diversity to when decoding with beam search.

\paragraph{BERT Pretraining} Large pretrained language models \cite{radford2019language,DBLP:conf/naacl/DevlinCLT19} have achieved impressive progress on NLP tasks \cite{DBLP:conf/emnlp/WangSMHLB18, DBLP:journals/corr/abs-2005-14165}. From Table \ref{tab:automatic}, we can see that BERT pretraining is also of great contribution for WPS. It suggests that the results might be further improved if BERT is pretrained on large-scale fiction corpora, which we leave as future work.

\paragraph{Insertion Bias} We observe that insertion bias also counts a lot. Injecting the bias to the decoder grants Locate\&Gen the control over insertion on arbitrary positions. Generation models without it would neglect the predicted insertion position and could not help producing identical similes for a given context. The improvement yielded by the bias show that in addition to the encoder-decoder attention mechanism, equipping the decoder with interpolation signals is also beneficial for generating coherent similes. 

\subsection{Case Study}
As shown in Table~\ref{tab:compare}, we perform case study to delve into the advantages and shortcomings of our Locate\&Gen model.

\paragraph{Making Meaningful Comparisons} For case \#1, all retrieval methods select similes containing the word ``cave'', resulting in similes as ``cavity is like a cave'', which fail to elicit aesthetic experiences. In comparison, both Locate\&Gen and ground-truth produce engaging similes that compare a cave with funnel and lid, respectively. 

\paragraph{Approaching Ground-Truth} All models perform well in case \#2 whereas the simile by MatchCBOW is a bit too specific given the current non-informative context. Note that Locate\&Gen yields the best similarity with ground-truth since they both apply the word ``freeze'' in similes to describe ``movements stopped abruptly''. 

\paragraph{Enhancing Drama} Case \#3 is even more interesting. Both ground-truth and Locate\&Gen generate not simply a simile, but an oxymoron (反衬) at the same time, which is another figure of speech where contradictory terms appear in conjunction. Specifically, ground-truth and Locate\&Gen use words like ``quail'' or ``mouse'' to describe ``Titans'', respectively. Moreover, ``mouse confronting a cat'' generated by Locate\&Gen is dramatic and coherent with the context ``Titans seeing their king''. 

\paragraph{Lacking Content Richness} For case \#4 however, we observe that retrieval methods especially MatchBERT outperform Locate\&Gen model in terms of generation diversity and richness. As a matter of fact, the overall performances across Table~\ref{tab:automatic} and Table~\ref{tab:compare} suggest the potentials of ensembling retrieval and generation methods.

\subsection{Inference Mode}
As introduced before, there are two inference modes available for WPS task. The studies on them are shown in Table~\ref{case:mode}. In automatic mode (boxed in yellow), the model firstly predicts the insertion position, and then generates a coherent simile. However, there are inevitably cases where the model's own simile interpolation is not as satisfying as expected, such as the simile ``like being hit by a train'', which seems a bit weird. In this case, we could thus change to semi-automatic mode (boxed in orange) 
where the model takes an extra input of desired insertion position and directly starts the biased simile generation. The design of insertion bias grants us the ability to control the simile interpolation at any desired position with contextual coherence.

\subsection{Shortcomings}
In spite of the interesting results, Locate\&Gen model still possesses several drawbacks. 

\paragraph{Lack of Explainability} on which target (a.k.a tenor (本体)) in the context is the predicted simile to describe. Take \#2 in Table~\ref{tab:compare} as example. The simile ``like being frozen'' is inserted after a comma, but it means to describe the target Zi Jiang's movement. Attention visualization might be indicative to some point, but it's not always feasible for interpretability \cite{DBLP:conf/acl/SerranoS19}. Since the insertion position and attention weights are not always aligned with the tenor, it remains an open question on how to explain the simile predicted by models to better assist human writings. One desirable way might be to first identify a target tenor as an additional input feature for Locate\&Gen model before 
performing simile interpolation. 

\paragraph{Lack of Diversity} and aestheticness of the generated simile, which is partly determined by the language complexity of training examples. As introduced before, we've distinguished the simile complexity with three levels, i.e., single, combined and clause as shown in Table~\ref{tab:stat}. Intuitively, similes with distinct levels of complexity accord with different styles of contents. Hence, another promising way to improve simile quality is to feed extra features of complexity and styles into the model. We believe it is a critical research problem to assist humans with simile polishment.

\paragraph{To Polish or Not To Polish} In real-world application, there are definitely cases where an input writing is already engaging enough and thus one does not need to gild the lily by adding more similes. As mentioned in the model section, our Locate\&Gen framework is actually compatible with such a scenario since there is a null pointer location that could be used to indicate unnecessary insertions, as long as proper negative training samples exist for the model to learn. Such case is however beyond the scope of this work and we plan to explore it in the future.


\section{Conclusion and Future Work}
In this paper, we introduce a new task, Writing Polishment with Similes, and curate a large-scale Chinese simile dataset. Our experiments demonstrate the feasibility and potential of the task, which we consider as a first step towards figurative writing polishment in a real-world setting. We establish Locate\&Gen model and benchmark it on the developed dataset.

Future works include but not limited to: 
\begin{itemize}
    \item Dataset refinement (e.g., neural simile recognition).
    \item Better model designs (e.g., retrieval-generation ensemble model has potentials, and tenor extraction as an additional feature).
    \item Focus on writing polishment tasks for other figurative types.
\end{itemize}

Furthermore, from an AI writing assistant perspective, we surmise that assisting humans with writing polishment is more likely to develop the potentials of current AI models than just letting AIs write on the fly (which has however become a typical slogan for cutting-edge generative models such as GPT3). Given that figurative language is an essential creative aspect of language use, we encourage the use of the CS dataset in various contexts and look forward to the emergence of intelligent writing assistant tools \textbf{like magic}\footnote[1]{We applied our Locate\&Gen model to generate this simile, which is ``如同魔术般的''(智能写作助手) in Chinese before being translated to English.} in the future. 

\bibliography{ref.bib}
\end{CJK*}
\end{document}